\documentclass{article}

\usepackage{arxiv}
\usepackage[utf8]{inputenc} 
\usepackage[T1]{fontenc}    
\usepackage[hidelinks]{hyperref}       
\usepackage{url}            
\usepackage{booktabs}       
\usepackage{amsfonts}       
\usepackage{nicefrac}       
\usepackage{microtype}      
\usepackage{lipsum}			
\usepackage{graphicx}
\usepackage{doi}
\usepackage{natbib} 
\usepackage{amsmath} 
\usepackage{booktabs} 
\usepackage{tikz}
\usepackage{amsmath}
\usepackage{algorithm}
\usepackage{algpseudocode}
\usepackage{bm} 
\usepackage{float}

\title{ENTROPY-REGULARIZED GRADIENT ESTIMATORS FOR APPROXIMATE BAYESIAN INFERENCE}

\author{ Jasmeet Kaur \\ 
	Department of Computer Science\\
	University of Texas, Austin\\
}

\date{}

\begin{document}

\maketitle

\begin{abstract}
Effective uncertainty quantification is important for training predictive models with limited data to improve accuracy and robustness. 
While Bayesian methods are effective for this purpose, they can be challenging to scale. 
Ensuring the quality of samples from the posterior distribution in a computationally efficient manner becomes essential when employing approximate Bayesian inference.
This paper addresses the estimation of the Bayesian posterior to generate diverse samples by approximating the gradient flow of the Kullback-Leibler (KL) divergence and the cross entropy of the target approximation under the metric induced by the Stein Operator. 
It presents empirical evaluations on classification tasks to assess the method's performance and discuss its effectiveness for Model-Based Reinforcement Learning that uses uncertainty-aware network dynamics models. 
\end{abstract}

\section{Introduction}\label{sec:intro}
 Deep Learning models often result in overconfident and miscalibrated predictions without uncertainty quantification.
 This is more pronounced in cases when limited data is available, the quality of data available is poor, or one needs to consider making out-of-distribution predictions. Moreover, such cases find their use in real-world situations that require safe decision-making.
Measuring uncertainty in predictions plays a crucial role in Reinforcement Learning (RL). Many works incorporate uncertainty to guide exploration, directing it toward regions of high uncertainty (e.g., \cite{lee2021sunrise}, \cite{liu2021cooperative}, \cite{ciosek2019better}, \cite{osband2016deep}).
\cite{chua2018deep}, \cite{fu2022model}, and \cite{janner2019trust} utilize ensembles of models to capture and represent uncertainty in learned environment dynamics, employing these models for planning purposes.
In many cases, deep ensembles effectively mitigate overestimation bias and stabilize Q-value functions that quantify the value of an action in a given state (\cite{chen2021randomized}).

Methods using ensembles have become popular for uncertainty modeling in deep neural networks. 
Ensembles combine several individual predictors. 
An ensemble's predictive performance depends on individual performance and the diversity among the individual members. 
Recent work,  \cite{abe2023pathologies}, \cite{abe2022best}, points out how the diversity of the members reduces as the networks get larger. 
In general, the effectiveness of uncertainty quantification using an ensemble of deep neural networks reduces as the predictive ability of the ensemble members increases. 
This can be attributed to having large enough in-distribution data, big ensemble size, or over-parameterized networks. Having training objectives that train an ensemble to make diverse predictions may seem like a possible solution, but this can harm the predictive performance (\cite{abe2023pathologies}).
This led to research to understand the cases under which ensembles are truly effective.
A good single model can outweigh the benefits of using an ensemble for making out-of-distribution predictions ( \cite{abe2022deep}).
However, diversity is still important for predictions in a limited data regime and can not be easily mitigated by increasing the model size without overfitting.

Recent work builds on the contributions of \cite{neal2012bayesian} and \cite{mackay1992bayesian}, extending Bayesian inference methods to develop predictive ensembles and improve uncertainty estimation.  
In Bayesian inference, one usually cares about the distribution of model parameters. Once this distribution is estimated, it can be used to make predictions by marginalizing. Bayesian Model Averaging is a way to do this in a tractable manner, with samples of this distribution constituting an ensemble.
Such a class of methods has been shown to improve generalization, enhance uncertainty quantification, and improve model reliability in handling out-of-distribution data.

Traditional approaches to Bayesian inference depend on obtaining samples from Markov Chain Monte Carlo (MCMC) chains. 
And, these become computationally inefficient for bigger networks. 
Some approximate methods, such as stochastic gradient MCMC methods, manage to generate samples of the posterior distribution with the help of gradient information and use mini-batches of data.
In comparison to these slow sampling methods, methods that directly optimize a tractable approximation of the true posterior tend to do better at inference time. These are often referred to as Explicit Variational Methods.
Implicit Variational methods, on the other hand, focus on directly estimating approximate posterior samples and have shown strong empirical performance 
(\cite{li2017gradient}). 

This paper considers a type of Implicit method: Particle optimization methods for approximate Bayesian Inference, which approximate the target distribution with an ensemble of networks acting as a set of particles and improve diversity among them. More specifically, this paper deals with the issue of \textit{mode collapse} that occurs when the different members of the ensemble make similar predictions. It introduces an approach to have functional diversity in posterior samples to avoid averaging over similar functions, which leads to redundancy in the model average 
(\cite{d2021repulsive}, \cite{wu2017bayesian}, \cite{pang2019improving}).
This phenomenon of mode collapse is more evident when networks are over-parameterized 
(\cite{d2021repulsive}, \cite{kirsch2024implicit})
and formally been described as the problem of non-identifiability (\cite{roeder2021linear}).

The work in this paper addresses the mode collapse issue and proposes a solution to ensure functional diversity using gradient information. It examines how functional diversity in particle optimization methods can be effectively controlled with kernel density estimation to approximate the target density. The paper introduces a simple method for training ensemble members for improved mode exploration and presents empirical results to validate the approach.

\subsection{Contribution}

In particular, the main contributions of this work are : 
\begin{itemize}
    \item A simple method to improve functional diversity by using kernel density as an approximation to the target distribution. 
    \item Show how this results from the gradient flow of KL divergence between the target distribution and its approximation and Cross-Entropy of this approximation with kernel density.
    \item Empirically study the implications of the method and test its efficacy in improving the planning for Model-Based Reinforcement Learning.
\end{itemize}

\section{Deep Ensembles using ERGD}
\subsection{Bayesian Neural Networks (BNNs)}
In a supervised learning setting, one has \( D = \{X,Y\}\) where \( X = \{ x_i \}_{i=1}^N \) denoting the training inputs and \( Y = \{ y_i \}_{i=1}^N \) denote the corresponding outputs, with \( x_i \in \mathcal{X} \) and \( y_i \in \mathcal{Y} \), respectively.

Let \( f(x, \theta): X \to \mathbb{R}^d \) denote a mapping function parameterized by a neural network, and  one can define a likelihood function using this 
\( p(y | f(x, \theta)) \).  

For example, a Gaussian for doing regression can be defined as : 
\[
p(y| f(x, \theta)) = \mathcal{N}(y|f(x; \theta), \sigma^2),
\]
where \( \sigma^2 \) is the variance of observation noise. 

For a classification problem with \( K \) classes, one could set \( d = K \) and let 
\[
p(y| f(x, \theta)) = \text{Multinomial}(y|\text{softmax}(f(x; \theta))).
\]

In Bayesian Neural Networks, the goal is to estimate the posterior distribution that captures all possible explanations for the given training data, defined as follows. 
\[
p(\theta|D) \propto  \prod_{i=1}^{N} p(y_{i}|x_{i},\theta) p(\theta)
\]
\( p(\theta) \) represents the prior distribution over the weight parameters \( \theta\). 

Most traditional ways to use neural network methods use a point estimate called a Maximum A Posteriori (MAP) estimate, which can be useful when sampling from a Bayesian Neural Network is challenging or when there is no need to capture model uncertainty. 

In this paper, the focus is on capturing a type of uncertainty called epistemic uncertainty. It is the uncertainty associated with estimating the model parameters given the training data. Epistemic uncertainty modeling is well-suited when data is limited or additional data collection is difficult. Although many works use epistemic uncertainty methods when distributional shifts are present, this could be potentially misleading. Ensembles are essential for improving predictive performance, uncertainty quantification, and robustness over a single model. However, their ability to quantify uncertainty and ensure robustness on out-of-distribution data correlates with their in-distribution performance. This falls outside the primary focus for discussion in this paper. But there are non-parametric ways that quantify uncertainty for such distribution shifts (\cite{malinin2018predictive}).

Finally, to predict a test data point ($x_{test}$, $y$),  one can marginalize the entire posterior distribution rather than relying on a single parameter estimate. 
\[
p(y_{\text{test}}|x_{\text{test}}, D) = \int p(y_{\text{test}}|x_{\text{test}}, w) p(dw|D)
\]
However, this integral is intractable for neural networks. In practice, it is approximated using Bayesian Model Averaging (BMA). Sampling-based methods for Bayesian Inference facilitate the acquisition of representative samples from such a posterior distribution to estimate this quantity. This is why prediction accuracy in most approximate methods relies heavily on this Bayesian Model Averaging estimate. 
\[
p(y_{\text{test}}|x_{\text{test}}, D) \approx \frac{1}{M}\sum_{i=1}^{M}  p(y_{\text{test}}|x_{\text{test}}, \theta_{i})
\]

\subsection{Particle-optimization Variational Inference}
This section describes how Implicit methods estimate the posterior as a stationary distribution by minimizing an energy functional. This stationary distribution results by simulating the system's dynamics for an infinite amount of time.
This can be modeled as a partial differential equation of the form :
\[
\frac{\partial q_t}{\partial t} = -\nabla \cdot (v \cdot q_t),
\]
where \( q_t \) is the approximate posterior at time \( t \). \( v \) is the gradient flow defined as moving in the direction of the steepest descent of an energy or objective function. This gradient flow depends on the choice of metric and energy functional. Samples obtained by simulating such a gradient flow constitute samples from the stationary distribution. In practice, numerical methods can be used for simulation.

Particle-optimization Variational Inference (POVI) methods approximate \( q_t \) with a set of particles \( \{ \theta^{(i)} \}_{i=1}^n \), i.e., 
\[
q_t(\theta) \approx \frac{1}{n} \sum_{i=1}^n \delta(\theta - \theta^{(i)}_t),
\]
 simulate the gradient flow with a discretized version of the ordinary differential equation (ODE):
\[
\frac{d\theta^{(i)}_t}{dt} = -v(\theta^{(i)}_t).
\]
The particles are updated using the following update in every iteration, 
\[
\theta^{(i)}_{t+1} \leftarrow \theta^{(i)}_t - \eta_t v(\theta^{(i)}_t),
\]
where \( \eta_t \) is a step size.

Implicit Variational methods, thus,  allow for the generation of samples without having to do probability evaluation for approximating the true posterior. This contrasts with Explicit variational methods, which approximate the posterior distribution by minimizing the KL divergence between a simpler, tractable distribution and the true posterior. The use of limited tractable distributions can hinder the accuracy of such methods.

Different POVI methods differ in their choices of metric spaces and functionals for gradient flows. 
A pivotal study that proved significant for subsequent research directions is Stein Variational Gradient Descent (SVGD) (\cite{liu2017stein}). \cite{d2021repulsive} shows how SVGD can be interpreted as a gradient flow in the Wasserstein Space under a particular Stein geometry.

\cite{jordan1998variational} described the optimization problem of a functional $F:  \mathcal{P}_2(\mathcal{M}) \to \mathbb{R}$ as the evolution in time of the measure $\rho$ under the Wasserstein gradient flow is described by 
{\small
\begin{equation}
\begin{split}
    \frac{\partial \rho(x)}{\partial t} &= \nabla \cdot \bigg(\rho(x) \nabla  \frac{\delta}{\delta \rho}F(\rho) \bigg). 
\end{split}
    \label{eqn:wgf}
\end{equation}
}\\
Here, $$\nabla  \frac{\delta}{\delta \rho}F(\rho)=: \nabla_{\mathcal{W}_2} F(\rho,\pi)$$ is the Wasserstein gradient with a target measure, and the operator $$\frac{\delta}{\delta \rho}: \mathcal{P}_2(\mathcal{M})  \to \mathbb{R}$$ represents the functional derivative.

For SVGD, the update it given by :

\begin{equation}
\begin{split}
   \frac{dx_{i}}{dt} = \nabla_{\mathbf{x}_{i}} \log \pi \left(\mathbf{x}_{i}\right) - \beta \sum_{j=1}^{k} \nabla_{\mathbf{x}_{i}} k \left(\mathbf{x}_{i}, \mathbf{x}_{j}\right)
\end{split}
\end{equation} 

In particular, if $\mathbb{K}_{\rho}(\phi)(x)$ is a linear operator defined as $\mathbb{E}_{x'\sim\rho}[k(x,x')\phi(x')]$ and kernel $k(x,x')$ captures the particle interactions (\cite{d2021repulsive}), the ODE describing the SVGD update becomes : 
\[
 \frac{dx_{i}}{dt} = \mathbb{K}_{\rho} \nabla_{x'}(\log\pi (x') - \log \rho(x'))
\]

Here, $\pi (x)$ is the true posterior and $\rho(x)$ is the target approximation.

The \textit{$p$-Wasserstein distance} for $W_2$ between $\rho_0$ and $\rho_1$ is  defines a metric on $\mathcal{P}_2$$(\mathbb{R}^d)$ as 
{\small
\begin{align}
W_2^2(\rho_0, \rho_1) &:= \inf_{Y_0 \sim \rho_0; \, Y_1 \sim \rho_1} \mathbb{E} \left( \| Y_0 - Y_1 \|^2 \right) \\
&= \inf_{\gamma \in \Pi(\rho_0, \rho_1)} \int_{\mathbb{R}^d \times \mathbb{R}^d} \| y_0 - y_1 \|^2 \, d\gamma(y_0, y_1).
\end{align}
}
Think of this as an optimal transport problem where one wants to transform elements in the domain of $\rho_0$ to $\rho_1$ with a minimum cost. 

\cite{d2021repulsive} introduces a method that uses Wasserstein gradient flow with kernel density to approximate the target distribution. In their proposed method, the functional is the KL divergence between the target approximation and the true posterior. The update for particle simulation in this work is as follows : 
\begin{equation}
\begin{split}
   \frac{dx_{i}}{dt} = \nabla_{\mathbf{x}_{i}} \log \pi \left(\mathbf{x}_{i}\right) - \beta \sum_{j=1}^{k} \nabla_{\mathbf{x}_{i}} k \left(\mathbf{x}_{i}, \mathbf{x}_{j}\right)
\end{split}
\end{equation}

\cite{chen2018unified} presents a novel particle-optimization framework grounded in Wasserstein gradient flows to unify two prominent scalable Bayesian sampling methods: stochastic gradient Markov chain Monte Carlo (SG-MCMC) and Stein variational gradient descent (SVGD).

One of the factors impacting the diversity of samples in all the methods mentioned above, obtained at the end of the simulation, is the initial particle distribution. Additionally, as the networks get larger, these samples become less and less sensitive to this initial distribution during training and converge to a single mode. Empirical evidence and theoretical analysis \cite{zhuo2018message} have shown that particles in methods such as SVGD exhibit mode collapse. They converge to a limited number of local modes. It is natural to ask if there is a way to reduce this dependence on the initial distribution and enhance mode coverage for diversity. This is not a novel research question; several prior works have looked into finding ways to answer this question to address mode collapse.

The proposed method, called Entropy-Regularized Gradient Estimators, is a method to enhance mode coverage by mitigating the ensemble's dependence on the initial distribution while accounting for particle interactions within the ensemble. It uses an empirical measure of particles to approximate the target posterior.  This set of particles is taken from an initial distribution and is evolved using the following update :

\begin{equation}
\begin{split}
      \frac{dx_{i}}{dt} = \frac{1}{n} \sum_{j=1}^n \left( k(\mathbf{x}_i, \mathbf{x}_j) \nabla_{\mathbf{x}_j} \log \pi (\mathbf{x}_j) - \beta \nabla_{\mathbf{x}_i} k(\mathbf{x}_i, \mathbf{x}_j) \right)
\end{split}
\end{equation}

The full algorithm for Entropy-Regularized Gradient Descent is described in Algorithm 1. A symmetric kernel is used to model the interaction between the particles to train an ensemble. 
ERGD introduces a regularization constant that controls the effect of this interaction on functional diversity.
The key idea behind this is that using kernel density as an approximation to the target density, the constant regularizes cross-entropy between the kernel density and this approximation. Intuitively, the evolution of the particles is influenced more by their interactions with each other. 

Particle evolution equation:
\begin{equation}
\frac{d\bm{x}}{dt} = - \int k(\bm{x}, \bm{x'}) \left[ \nabla_{\bm{x'}} \frac{\delta \mathcal{F}(\rho)}{\delta \rho(\bm{x'})} \right] \rho(\bm{x'}) \, d\bm{x'}
\end{equation}
\noindent
\textit{where} \( \bm{x} \in \mathbb{R}^d \) is the particle,  
\( k(\bm{x}, \bm{x'}) \) is a symmetric positive-definite kernel, and  
\( \rho(\bm{x'}) \) is the particle density.

\vspace{1em}

Total energy functional:
\begin{equation}
\mathcal{F}(\rho) = (-1 + \beta) H(\rho, \kappa) + D_{KL}(\rho \,\|\, \pi)
\end{equation}
\noindent
\textit{where} \( \beta \in \mathbb{R} \) is a parameter,  
\( H(\rho, \kappa) \) is the cross-entropy between particle density $\rho$ and kernel density $\kappa$, 
\( D_{KL} \) is the Kullback-Leibler divergence between \( \rho \) and target distribution \( \pi \).
\begin{algorithm}
\caption{Entropy-Regularized Gradient Descent (ERGD)}
\begin{algorithmic}[1]
\State \textbf{Input:} Particles $\{x_1, x_2, \dots, x_n\}$, kernel function $k$, target posterior $\pi(x)$, regularization constant $\beta$
\For{each particle $x_i$}
    \State Compute the gradient $\nabla_{x_j} \log \pi(x_j)$ for all particles $x_j$
    \State Update the particle using the following rule:
    \[
    \frac{dx_i}{dt} = \frac{1}{n} \sum_{j=1}^{n} k(x_i, x_j) \nabla_{x_j} \log \pi(x_j) - \beta \nabla_{x_i} k(x_i, x_j)
    \]
\EndFor
\State \textbf{Output:} Updated particles $\{x_1, x_2, \dots, x_n\}$
\end{algorithmic}
\end{algorithm}
\vspace{1em}

This update is similar to the SVGD update except for the regularization term.
The evolution of the particles depends upon two types of updates. The first part on the right side considers the interaction between the particles as the attraction between them, and the second term is the repulsive term. Any symmetric kernel from the Stein Class can be used as an approximation to the target density.

The convergence guarantees of the method determine the choice of this regularization constant. As $\lim_{t\to\infty}  \beta \rightarrow 1$, $\rho = \pi$ is a solution of the corresponding Louiville Equation (more details in Appendix A). A simple choice for $\beta$ satisfying this is to use a linear schedule between positive values bigger than 1 to 1. This provides a method to guide particle evolution, placing much greater emphasis on the interactions between the particles in the initial distribution.

Using a symmetric kernel as an approximation to the target probability density changes the gradient flow of KL divergence between two distributions under the metric induced by the Stein Operator to the gradient flow of KL divergence under the Wasserstein Metric (\cite{d2021repulsive}).
The following update gives the particle evolution method described in \cite{d2021repulsive} : 
\begin{equation}
\begin{split}
   \frac{dx_{i}}{dt} = \nabla_{\mathbf{x}_{i}} \log \pi \left(\mathbf{x}_{i}\right) - \beta \sum_{j=1}^{k} \nabla_{\mathbf{x}_{i}} k \left(\mathbf{x}_{i}, \mathbf{x}_{j}\right)
\end{split}
\end{equation}

It can be understood as having the ensemble members grow without attraction, leading to a high probability density. In this case, the authors used $\beta$ as 1 / $\sum_{j=1}^{k} k \left(\mathbf{x}_{i}, \mathbf{x}_{j}\right)$.  
Such a $\beta$ can be bad in practice (\cite{d2021repulsive}, \cite{d2021repulsive}). 
\cite{kirsch2024implicit} developed gradient estimators by inverting Stein’s Identity. The work demonstrates that their method of selecting $\beta$ outperforms the gradient estimator proposed in \cite{d2021repulsive}, as their estimator accounts for the effects of all interactions on a particle, including those not involving the particle itself. 
ERGD can direct particle evolution based on an initial distribution, but it's performance depends on this initial distribution. To find a distribution that optimizes for mode coverage, ERGD is extended to $s$-ERGD as follows : 

\begin{equation}
\begin{split}
   \frac{dx_{i}}{dt} = \nabla_{\mathbf{x}_{i}} \log \pi \left(\mathbf{x}_{i}\right) - \frac{\beta}{n} \sum_{j=1}^{k} \nabla_{\mathbf{x}_{i}} k \left(\mathbf{x}_{i}, \mathbf{x}_{j} \right)
\end{split}
\end{equation} $s$-ERGD update makes $\beta$ a tunable parameter in the algorithm.
Experiments comparing different updates use $\beta > 1$ for $s$-ERGD. 

\section{Experiments}
This section describes experiments to compare ERGD and $s$-ERGD with other POVI methods and Deep Ensembles. 
\subsection{Bivariate Gaussian Mixture}
The first set of experiments tests how well different methods fit Gaussian mixture model. The centers for these Gaussians are at (213, 200), (180, 200), (200, 210), (200,190) with weights 0.6, 0.3, 0.05, 0.05. The experiments use 300 particles initialized at (180, 180).

It is important to note that s-ERGD may fail to converge in certain cases. Using ERGD with a linear schedule still fails to capture all modes of the distribution. \cite{d2021annealed} suggests that a linear schedule could result in slower training dynamics and proposed the use of a cyclic tanh schedule as an alternative. While ERGD with a tanh schedule can capture some of the modes, it does not necessarily guarantee complete mode discovery. On the other hand, s-ERGD is capable of exploring the distribution more effectively. Therefore, in the second-to-last figure of the final row, s-ERGD is employed to obtain an initial distribution, which is subsequently converges to the final solution using a linear schedule. This approach allows s-ERGD to serve as a way to have an initial distribution, while having the convergence properties associated with ERGD.

\begin{figure}[htbp]
    \centering
    \begin{minipage}{0.18\columnwidth} 
        \centering
        \includegraphics[width=\textwidth]{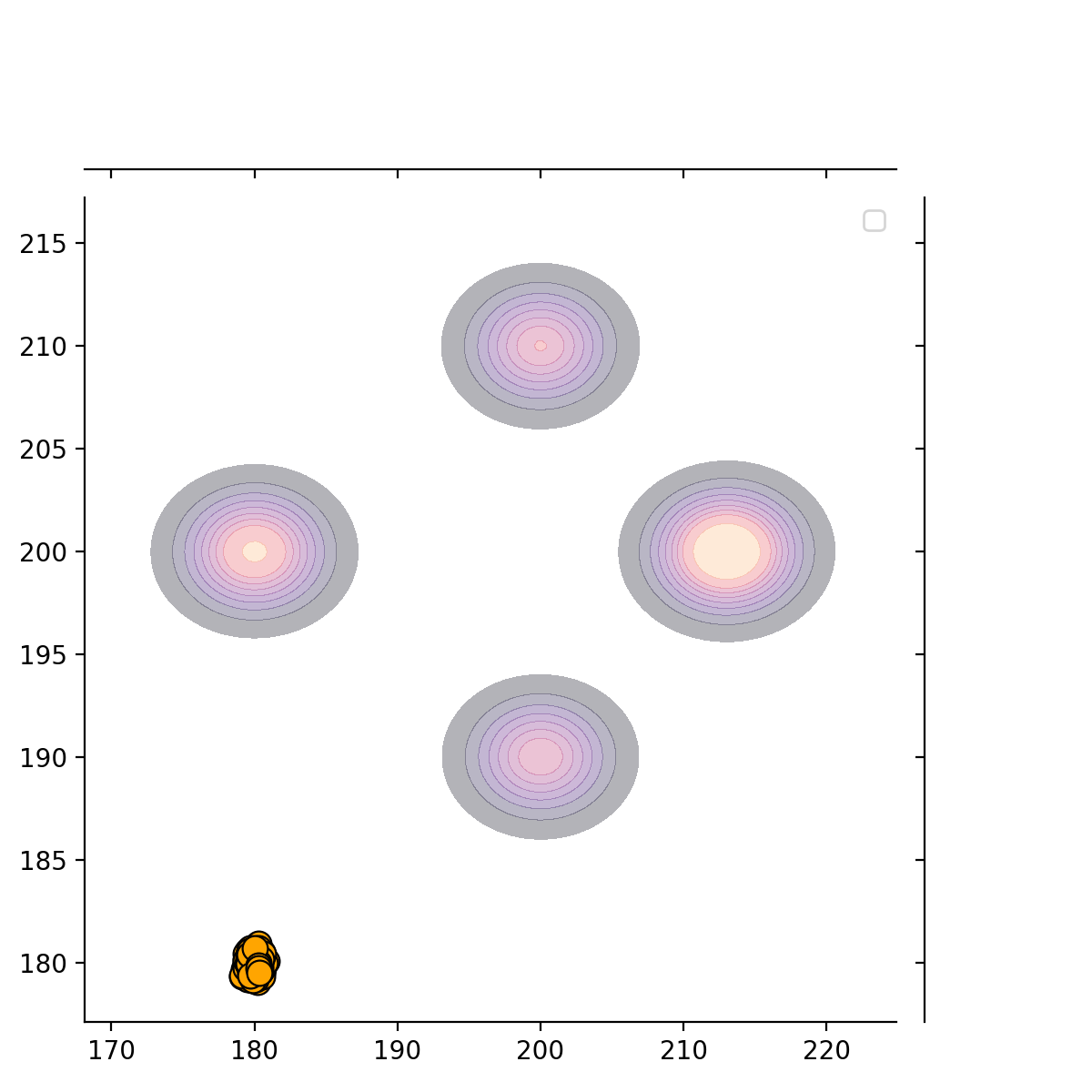}
    \end{minipage}\hfill
    \begin{minipage}{0.18\columnwidth}
        \centering
        \includegraphics[width=\textwidth]{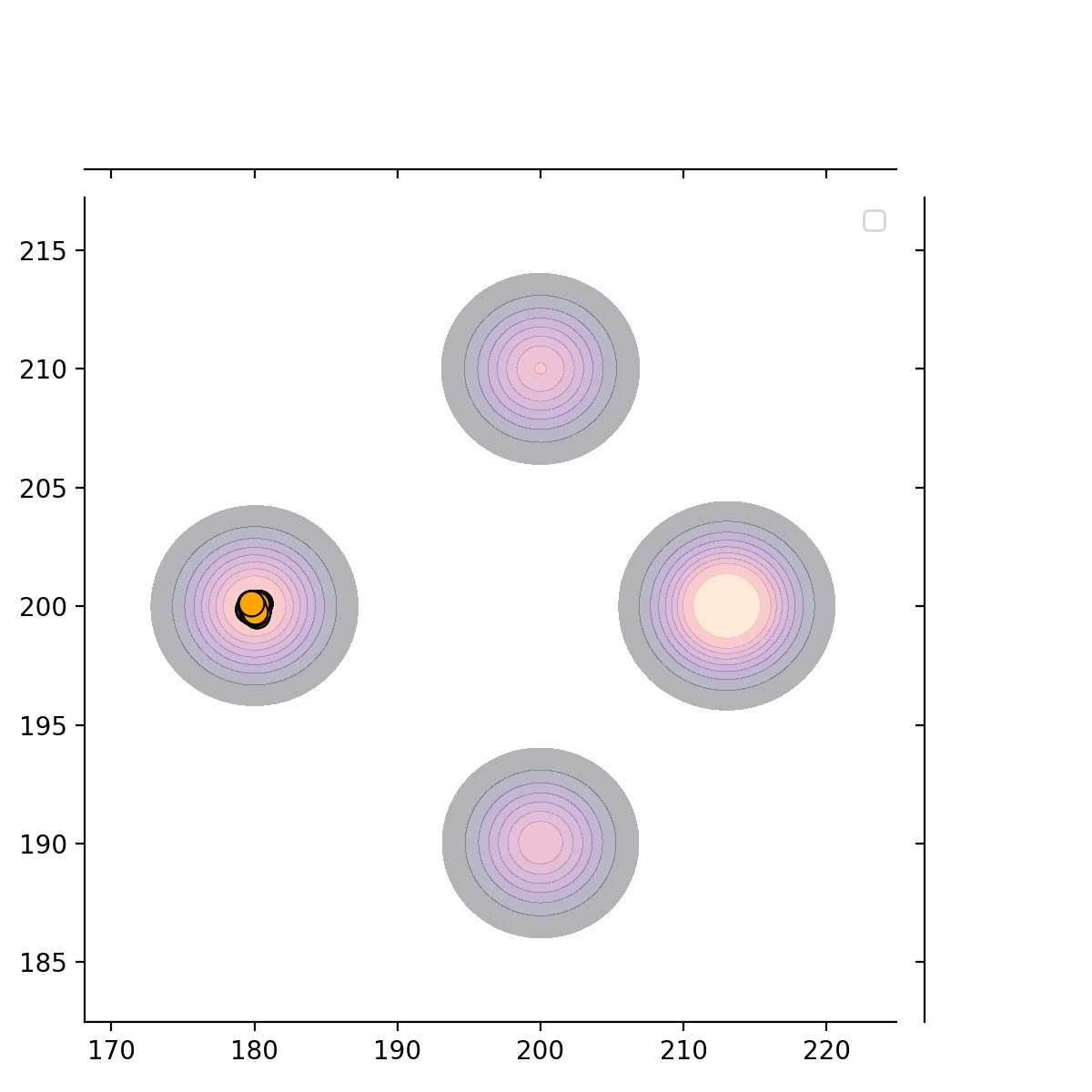}
    \end{minipage}\hfill
    \begin{minipage}{0.18\columnwidth}
        \centering
        \includegraphics[width=\textwidth]{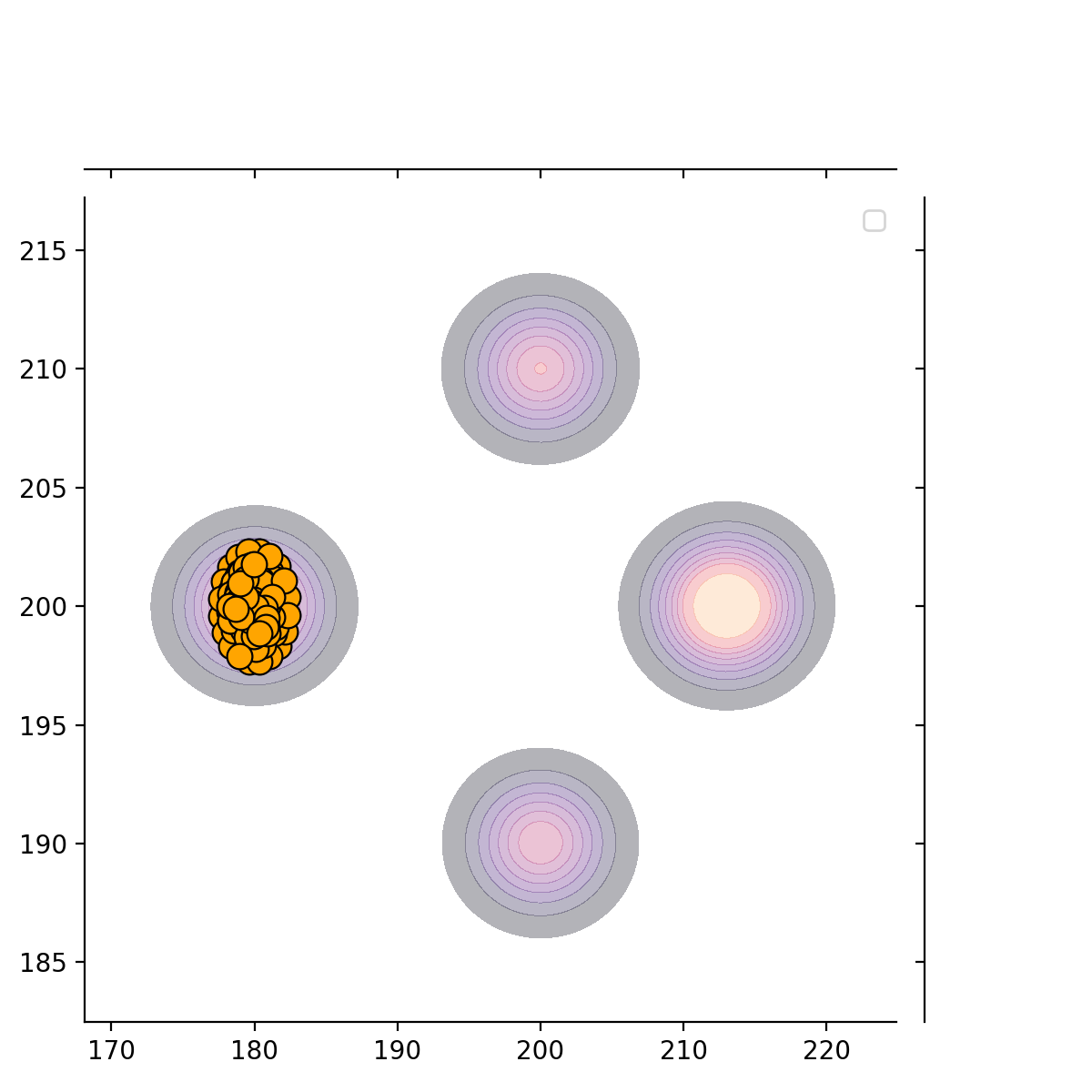}
    \end{minipage}\hfill
    \begin{minipage}{0.18\columnwidth}
        \centering
        \includegraphics[width=\textwidth]{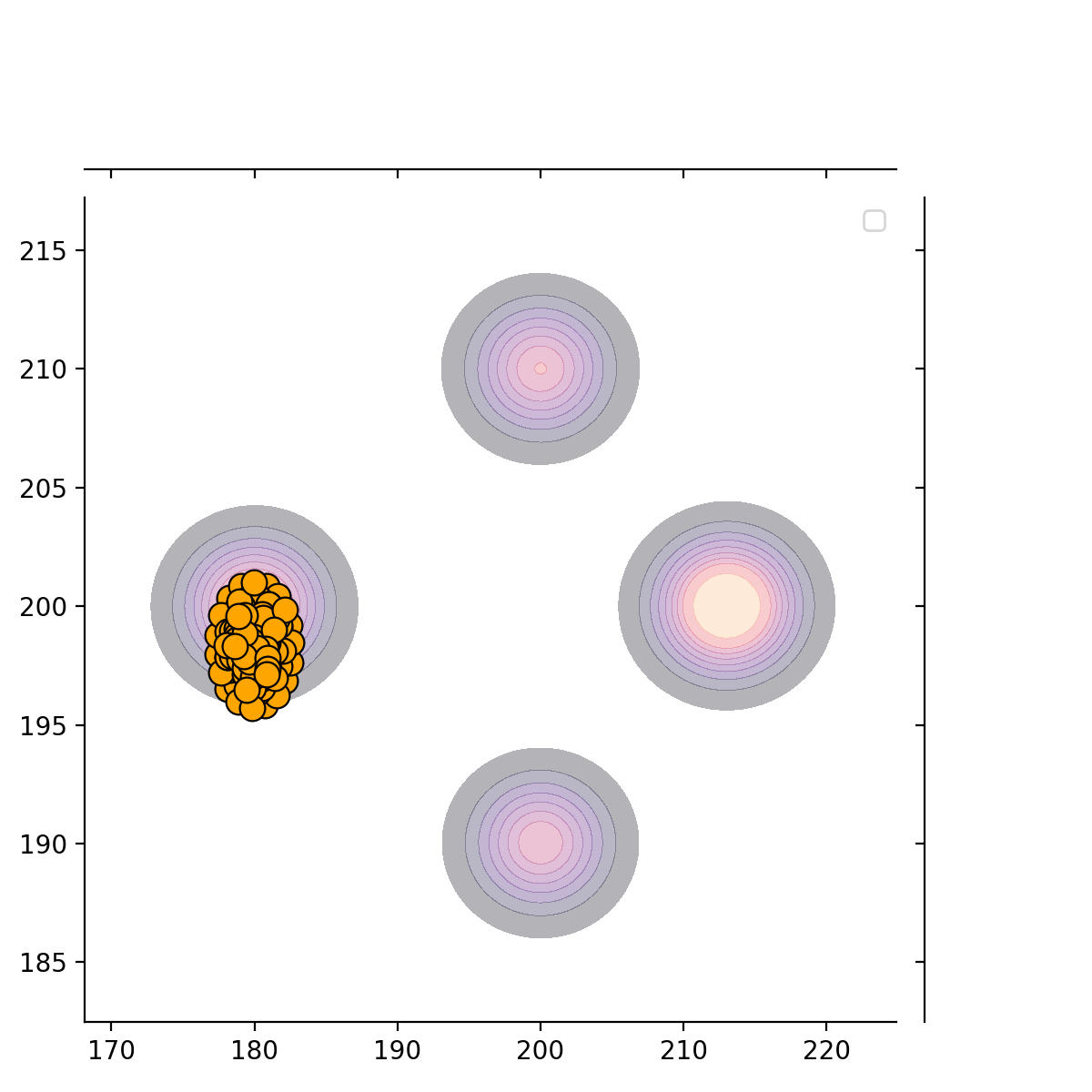}
    \end{minipage}\hfill
    \begin{minipage}{0.18\columnwidth}
        \centering
        \includegraphics[width=\textwidth]{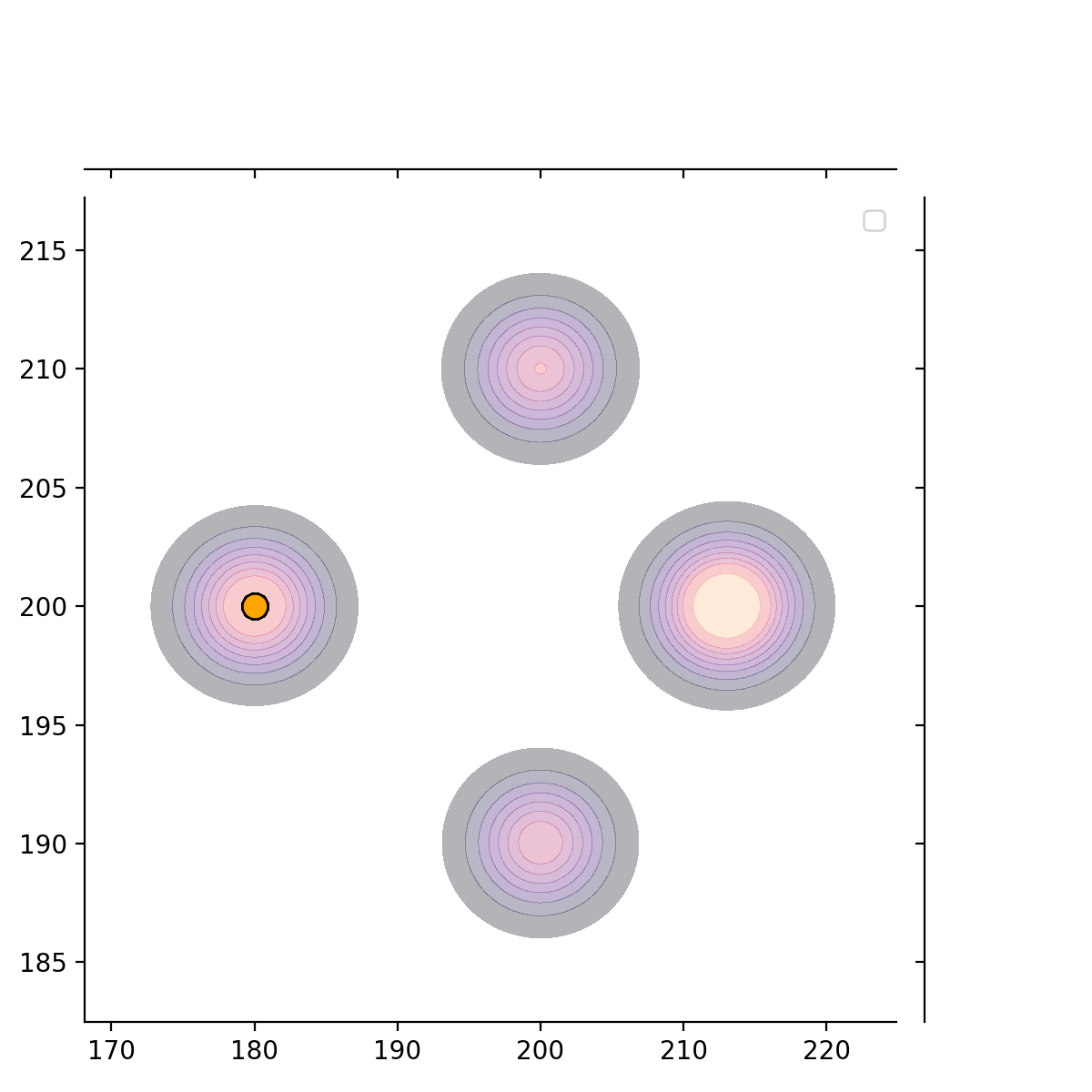}
    \end{minipage}
    
    \vskip\baselineskip 
    \begin{minipage}{0.18\columnwidth} 
        \centering
        \includegraphics[width=\textwidth]{imgs/random23_niter1500_init_anneal_500_beta_180_lr0.5_particles300_train10000Init_Gaussians.png}
    \end{minipage}\hfill
    \begin{minipage}{0.18\columnwidth} 
        \centering
        \includegraphics[width=\textwidth]{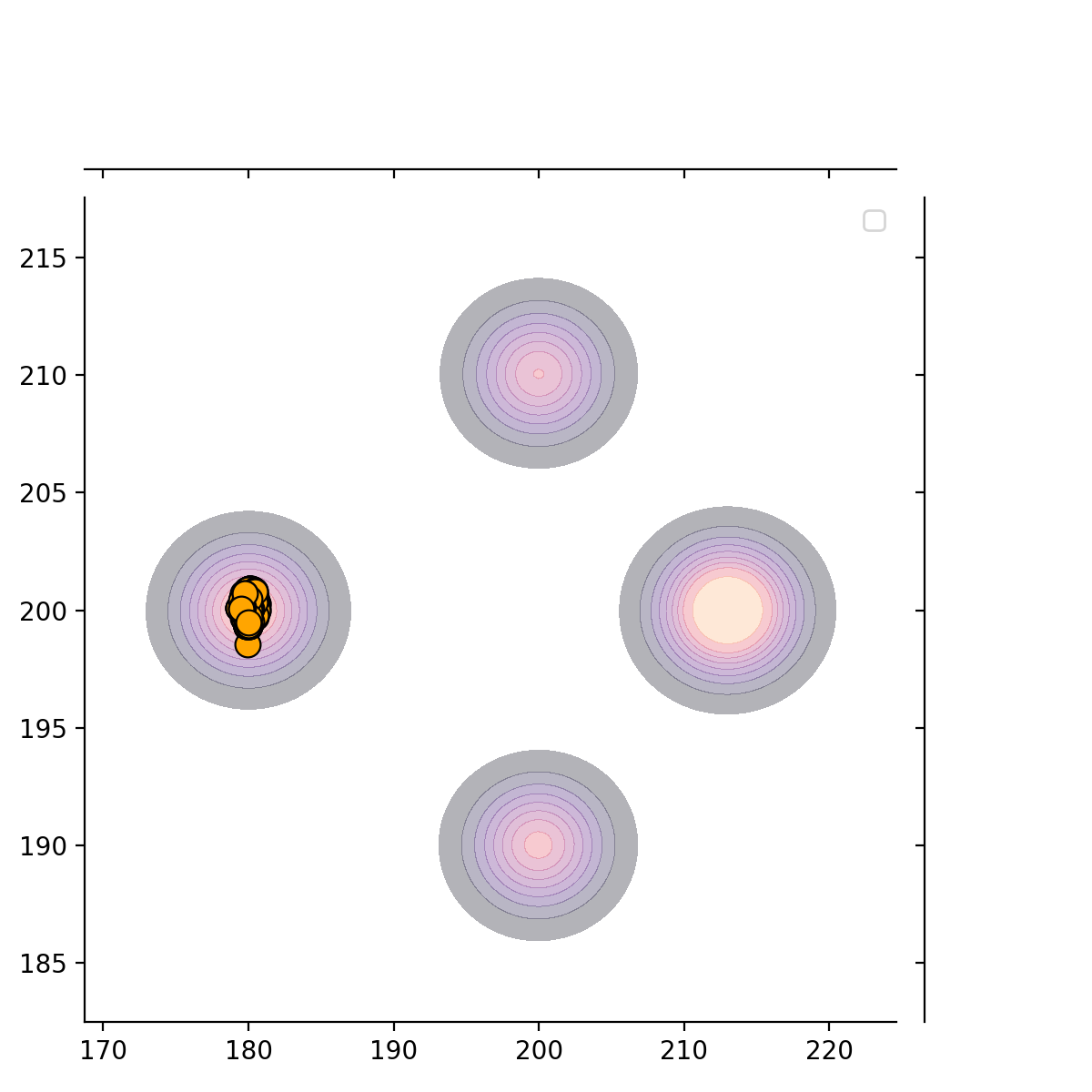}
    \end{minipage}\hfill
    \begin{minipage}{0.18\columnwidth}
        \centering
        \includegraphics[width=\textwidth]{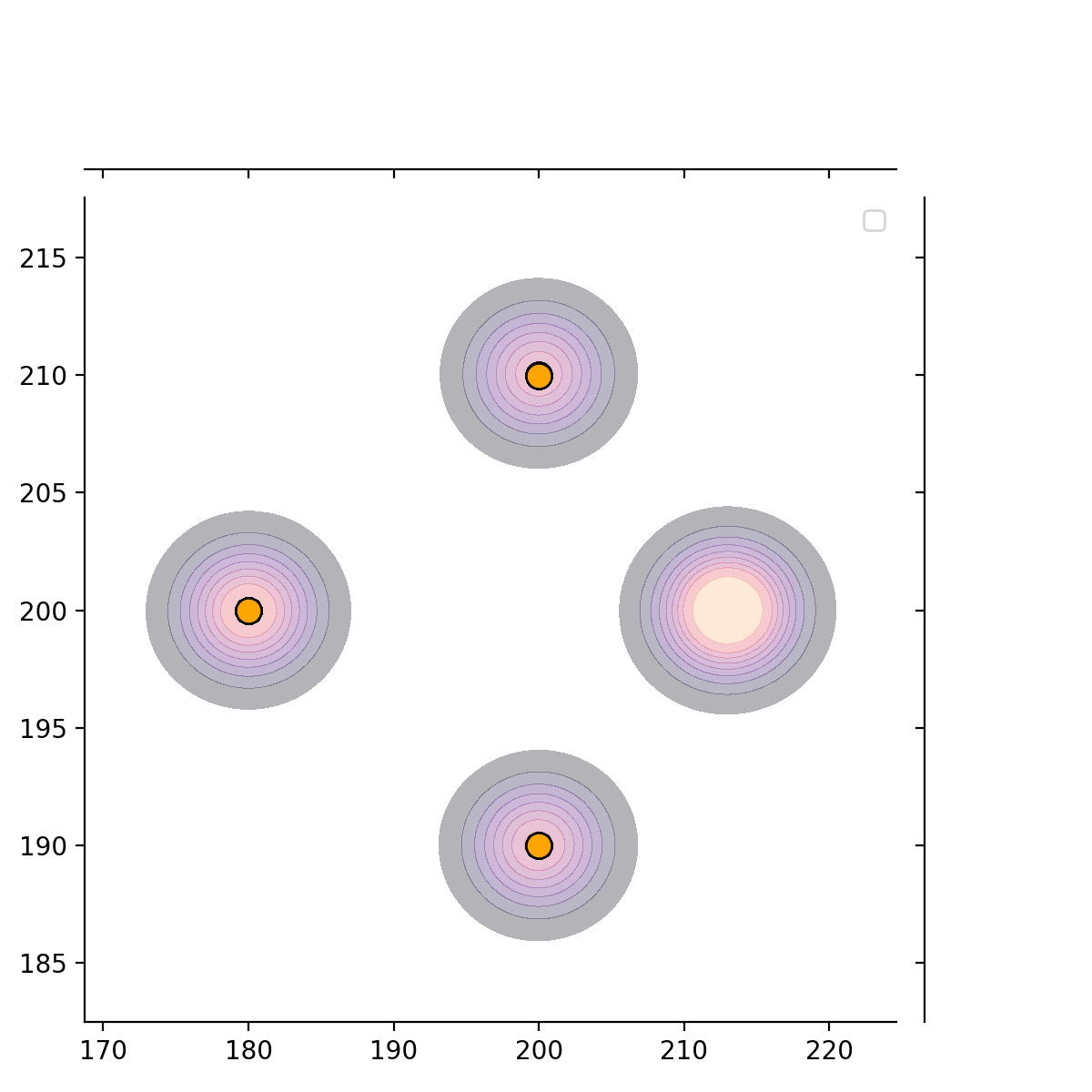}
    \end{minipage}\hfill
    \begin{minipage}{0.18\columnwidth}
        \centering
        \includegraphics[width=\textwidth]{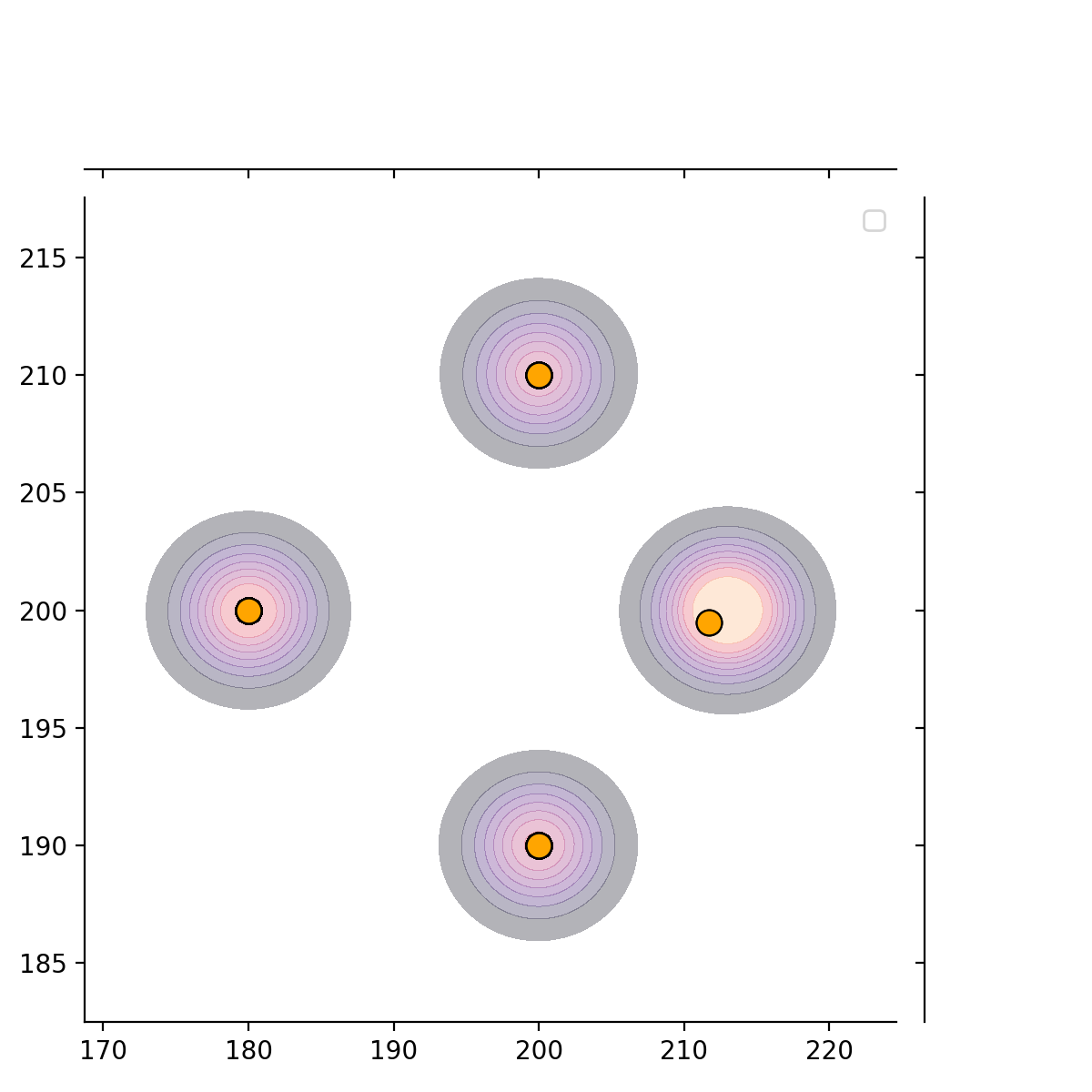}
    \end{minipage}\hfill
    \begin{minipage}{0.18\columnwidth}
        \centering
        \includegraphics[width=\textwidth]{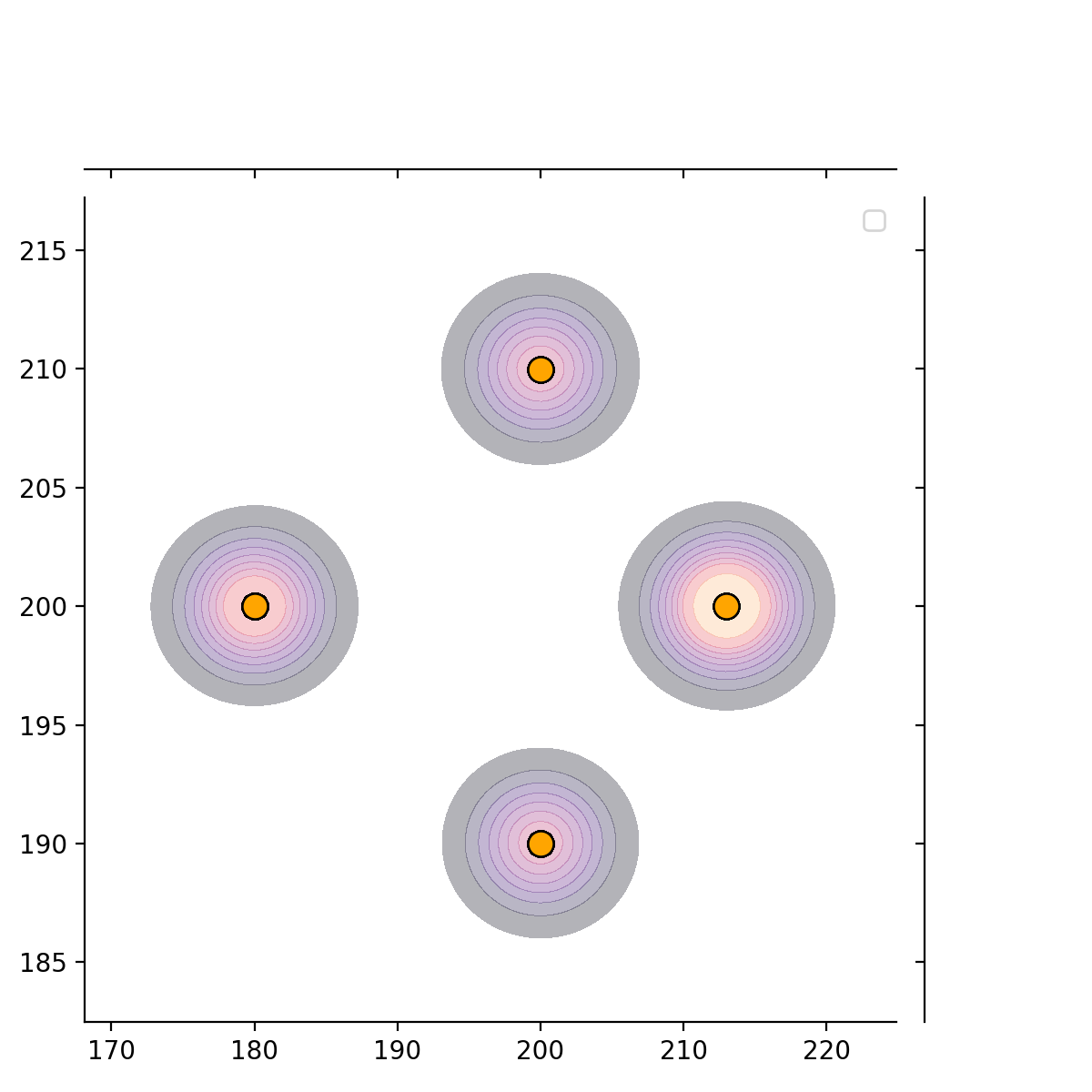}
    \end{minipage}
    
    \caption{Starting from top row left - Initial Distribution, SVGD, kde-WGD, sse-WGD, SGD. Initial Distribution, ERGD with a linear and tanh schedule, ERGD with $s$-ERGD for an initial Distribution followed by a linear schedule and $s$-ERGD.}
\end{figure}

\subsection{Classification Tasks}

The classification experiments are conducted on the CIFAR-10 dataset and Fashion-MNIST dataset. The results report the performance of all the methods in terms of accuracy on the test data and negative-log-likelihood. The ratio between predictive entropy on OOD and test data points (Ho/Ht) and the model disagreement (MD) ratio tests the diversity of the ensemble generated by the different methods. The results are averaged over 5 random seeds. SVHN dataset is used as the OOD benchmark for CIFAR-10 and MNIST dataset as the OOD benchmark for Fashion-MNIST.

Different methods are compared for an ensemble size of 10 and 20 for the classification of CIFAR-10 and FMNIST, respectively. Classification on CIFAR-10 is done using ResNet 20 and Fashion-MNIST uses a neural network with 3 hidden layers and 128 hidden units. ERGD and $s$-ERGD are compared with SGD, \textit{kde}-WGD using an update rule given by [6], \textit{sse}-WGD as described in \cite{kirsch2024implicit}, and SVGD. ERGD uses a linear annealing schedule. For $s$-ERGD, experiments use a constant value of beta to compare with \textit{kde}-WGD and \textit{sse}-WGD.

ERGD performs better than SVGD and is comparable to \textit{kde}-WGD  and \textit{sse}-WGD for both datasets. Deep Ensembles perform comparably to \textit{kde}-WGD on CIFAR-10 and have comparative performance to $s$-ERGD on FMINST.

\renewcommand{\arraystretch}{1.2}
\begin{table}[h!]
    \centering
    \scriptsize
    \begin{tabular}{p{2.4cm} p{2.4cm} p{2.4cm}}
        \toprule
         \textbf{Method} & \textbf{Accuracy} & \textbf{NLL} \\ 
        \midrule
        Deep Ensembles & $85.93 \pm 2.55$ & $0.29 \pm 0.02$ \\ 
        SVGD & $85.35 \pm 1.17$ & $0.28 \pm 0.03$ \\ 
        $kde$WGD & $85.625 \pm 0.891 $ & $0.274 \pm 0.02$ \\ 
        $sse$WGD & $85.979 \pm 1.06$ & $0.279 \pm 0.02$ \\ 
        ERGD & $85.625 \pm 3.68 $ & $0.28 \pm 0.03$ \\ 
        $s$ERGD & $83.75 \pm 2.61$ & $0.32 \pm 0.03$ \\ 
        \bottomrule \\ 
    \end{tabular}
    \hspace{1cm} 
    \begin{tabular}{p{2.4cm} p{2.4cm} p{2.4cm}}
        \toprule
        \textbf{Method} & \textbf{H$_{o}$/H$_{t}$} & \textbf{MD$_{o}$/MD$_{t}$} \\ 
        \midrule
        Deep Ensembles & $2.303 \pm 0.12$ & $1.878 \pm 0.16$ \\ 
        SVGD & $2.211 \pm 0.28$ & $1.954 \pm 0.25$ \\ 
        $kde$WGD & $2.319 \pm 0.38$ & $1.97 \pm 0.31$ \\ 
        $sse$WGD & $2.152 \pm 0.25$ & $1.891 \pm 0.26$ \\ 
        ERGD & $2.216 \pm 0.3$ & $1.927 \pm 0.23$ \\ 
        $s$ERGD & $2.111 \pm 0.192$ & $1.74 \pm 0.22$ \\ 
        \bottomrule \\
    \end{tabular}
    \caption{\textit{BNN Image Classification on CIFAR 10.}}
    \label{tab:split_table}
\end{table}

\vspace{0.3cm} 

\renewcommand{\arraystretch}{1.2}
\begin{table}[h!]
    \centering
    \scriptsize
    \begin{tabular}{p{2.4cm} p{2.4cm} p{2.4cm}}
        \toprule
        \textbf{Method} & \textbf{Accuracy} & \textbf{NLL} \\ 
        \midrule
        Deep Ensembles & $89.296 \pm 2.13$ & $0.179 \pm 0.02$ \\ 
        SVGD & $90.859 \pm 1.671 $ & $0.13 \pm 0.01$ \\ 
        $kde$WGD & $91.33 \pm 1.57$ & $0.127 \pm 0.017$ \\ 
        $sse$WGD & $91.016 \pm 1.63$ & $0.128 \pm 0.02$ \\ 
        ERGD & $91.016 \pm 1.56$ & $0.127 \pm 0.02$ \\ 
        $s$ERGD & $89.531 \pm 1.596$ & $0.182 \pm 0.02$ \\ 
        \bottomrule \\ 
    \end{tabular}
    \hspace{1cm} 
    \begin{tabular}{p{2.4cm} p{2.4cm} p{2.4cm}}
        \toprule
        \textbf{Method} & \textbf{H$_{o}$/H$_{t}$} & \textbf{MD$_{o}$/MD$_{t}$} \\ 
        \midrule
        Deep Ensembles & $4.382 \pm 0.57$ & $5.983 \pm 0.637$ \\ 
        SVGD & $5.819 \pm 1.11$ & $6.218 \pm 1.05$ \\ 
        $kde$WGD & $5.96 \pm 1.138$ & $6.31 \pm 1.16$ \\ 
        $sse$WGD & $5.884 \pm 1.195$ & $6.231 \pm 1.164$ \\ 
        ERGD & $5.67 \pm 0.921$ & $6.12 \pm 0.86$ \\ 
        $s$ERGD & $4.331 \pm 0.53$ & $5.984 \pm 0.631$ \\ 
        \bottomrule \\ 
    \end{tabular}
    \caption{\textit{BNN Image Classification on FMNIST.}}
    \label{tab:split_fmnist}
\end{table}

\begin{figure}[htbp]
    \centering
    \begin{minipage}{0.32\columnwidth} 
        \centering
        \includegraphics[width=\textwidth]{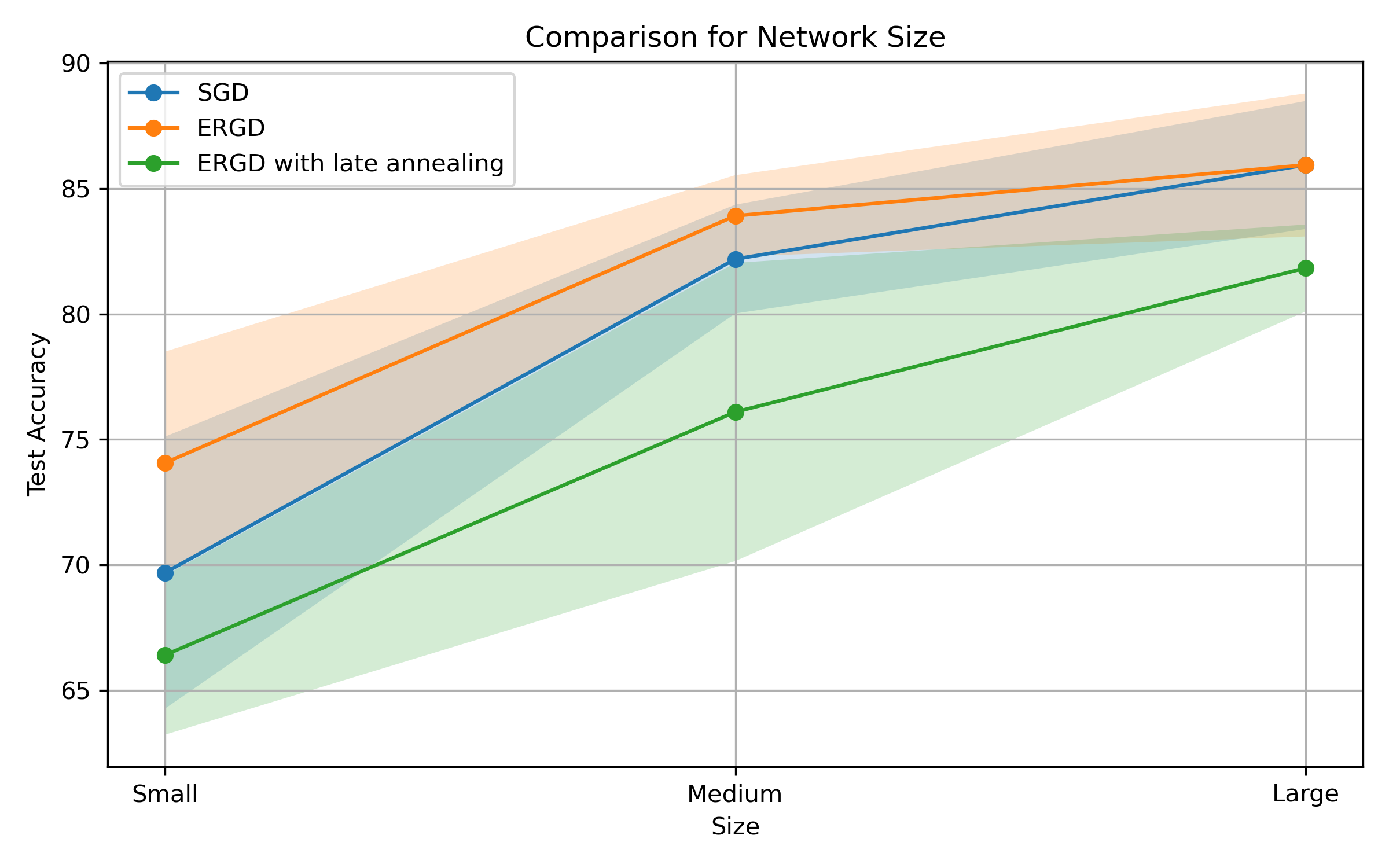}
    \end{minipage}\hfill
    \begin{minipage}{0.32\columnwidth}
        \centering
        \includegraphics[width=\textwidth]{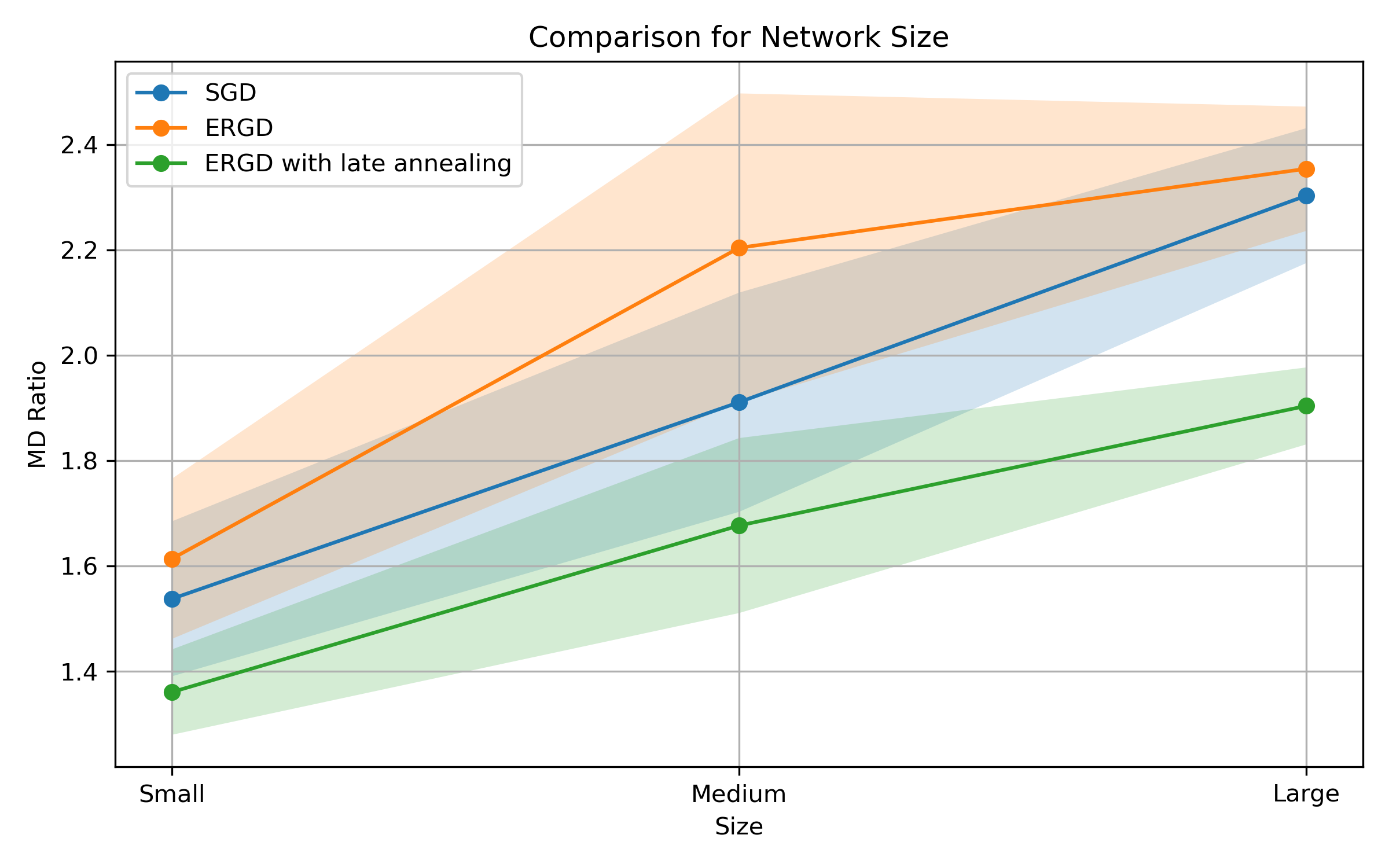}
    \end{minipage}\hfill
    \begin{minipage}{0.32\columnwidth}
        \centering
        \includegraphics[width=\textwidth]{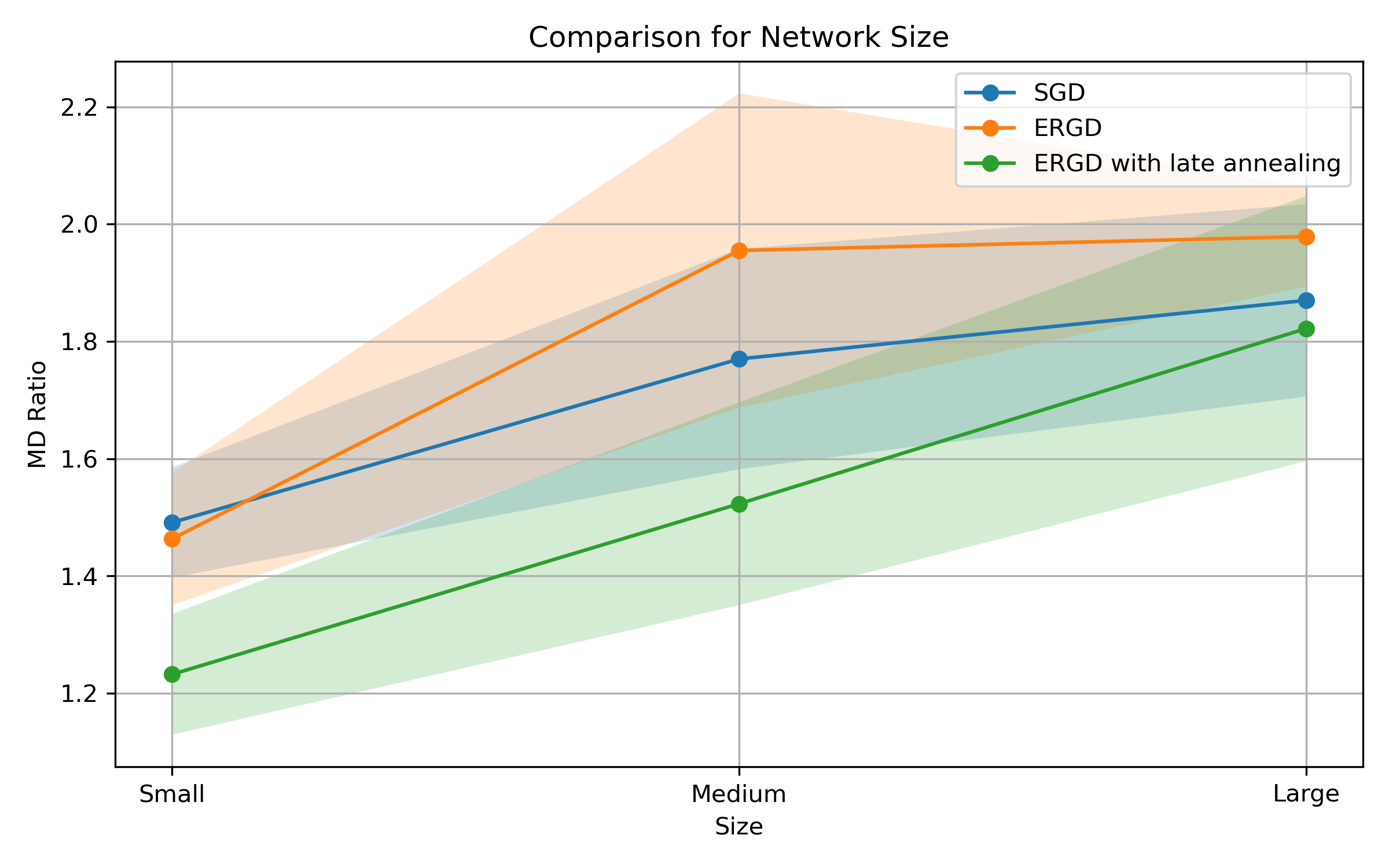}
    \end{minipage}\hfill

    \caption{Comparing SGD with ERGD for Network Size}
\end{figure}
Additionally, the performance of ERGD is compared with SGD as the network size changes. For this, the CIFAR-10 classification task is used, and the size of the ResNet is changed. There are different sizes compared, while the size of the ensemble is the same, i.e., 5. The green plot describes using ERGD with an initial exploratory phase, which does not add any advantage over ERGD for this case. ERGD has higher values of Entropy Ratio and Model disagreement compared to SGD in all cases, while the test accuracy gets close as the size increases.

\subsection{Uncertainty-aware Dynamics Models} 

This section presents experiments for model-based RL to test how well an ensemble's diverse predictions can be utilized to learn well when limited data is available. 
The section tests its efficacy in planning using uncertainty-aware model dynamics.
The experiments use the algorithm probabilistic ensembles with trajectory sampling (PETS) proposed in \cite{chua2018deep}.
PETS selects actions using model-predictive control. It plans for short horizons and optimizes for this by using ensembles to make model predictions for different action sequences.  
It solves tasks such as continuous cartpole, half cheetah, pusher, and reacher in less than 100,000 environment interactions.
It has been shown to perform more efficiently than SAC, PPO, or DDPG.

Experiments compare different ensembling methods with Deep Ensembles used in PETS. The ensembles are not trained from scratch at the start of every new episode. At the end of every episode, the converged target posterior informs training for the next episode. 
These experiments use an ensemble of 5 members and show results for all different optimization methods described above compared with SGD, which is used in PETS.
Currently, this has been done for three environments: continuous cart pole, half-cheetah, and pusher. The planning horizon for both half-cheetah and pusher is 30, and that for the cart pole is 25. Experiments are average episode rewards over 400 episodes of length 1000 for Halfcheetah, 50 episodes of length 100 for Pusher and 50 episodes of length 200 for Cartpole. Results in Table 3 are mean over only 2 runs for Halcheetah and Pusher and 3 runs for Carpole. 

\begin{figure}
    \centering
    \begin{tikzpicture}
    \node[anchor=north west] (img1) at (0, 0) {\includegraphics[width=2.5cm, height=2.5cm]{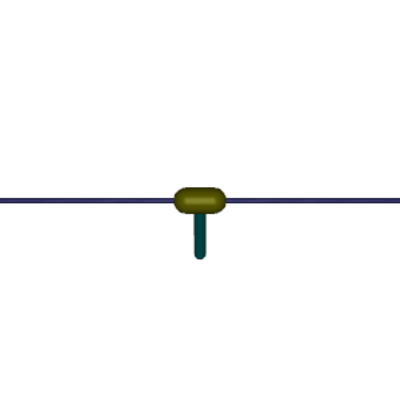}};
    \node[anchor=north west] (img2) at (3, 0) {\includegraphics[width=2.5cm, height=2.5cm]{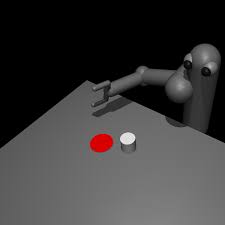}};
    \node[anchor=north west] (img3) at (6, 0) {\includegraphics[width=2.5cm, height=2.5cm]{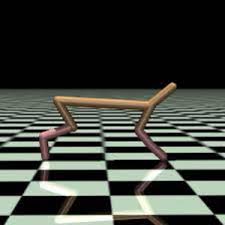}};
    \end{tikzpicture}
    \caption{Mujoco Environments [\cite{todorov2012mujoco}]} 
    \label{fig:enter-label}
\end{figure}

\renewcommand{\arraystretch}{1.2}
\begin{table}[h!]
    \centering
    \scriptsize
    \begin{tabular}{p{1.8cm} p{1.4cm} p{1.4cm} p{1.4cm}}
        \toprule
         \textbf{Method} & \textbf{Cartpole} & \textbf{Pusher} &\textbf{Halfcheetah} \\ 
        \midrule
        Deep Ensembles & $195.74 \pm 0.73$ & $-68.99 \pm 0.99$ & $5393.68 \pm 1441.12$\\ 
        SVGD & $197.25 \pm 2.11$ & $-53.321 \pm 1.28$ & $3599.64 \pm 292.95$ \\ 
        $kde$WGD & $196.94 \pm 0.29 $ & $-45.38 \pm 3.15$ & $5178.73 \pm 644.01$\\ 
        $sse$WGD & $196.48 \pm 1.04$ & $-50.81 \pm 2.3$ & $4984.58 \pm 807.59$\\ 
        ERGD & $196.94 \pm 0.45 $ & $-49.48 \pm 0.58$ & $6184.95 \pm 908.18$ \\ 
        $s$ERGD & $196.02 \pm 0.56$ & $-62.13 \pm 9.415$ & $4449.11 \pm 441.31$\\ 
        \bottomrule \\ 
    \end{tabular}
    \caption{\textit{Average Episode Reward for Mujoco Environments.}}
\end{table}

\section{Related Work}

Deep Ensembles are an effective way of quantifying uncertainty while naively maintaining functional diversity with no strong guarantee (\cite{wilson2020bayesian} \cite{izmailov2018averaging}). This is theoretically motivated by using constant SGD as an approximate Bayesian posterior inference algorithm (\cite{stephan2017stochastic}). \cite{maddox2019simple} builds on the idea of Stochastic Weight Averaging. It estimates a Gaussian approximation using stochastic gradient descent (SGD) with a constant learning rate and averaging the weights throughout training. \cite{wilson2020bayesian} integrate this concept with an underlying ensemble structure based on multiple runs of SWAG. Finding parameter-efficient ensembles has been considered by \cite{gal2016dropout}, \cite{srivastava2014dropout}, and \cite{maddox2019simple}. \cite{maddox2019simple} tackles the issue of scalability and efficiency of Bayesian Neural Networks using rank-1 approximation to the weight matrices.

The significance of training ensembles to produce diverse predictions has long been a standard approach.
However, recent studies have gone in depth in understanding the independent effect of using ensembles.
Some works suggest that increasing diversity during training often leads to less effective ensembles. 
\cite{abe2022best} views it from Jensen's gap. \cite{theisen2024ensembles} indicates that as long as the disagreement rate exceeds the average error rate, the ensemble continues to exhibit robust performance. Given extensive data, ensembling large networks does not prove more effective because it is not as sensitive to random initialization (\cite{abe2023pathologies}). Ensembling smaller networks works better. Works such as \cite{mania2020classifier}, \cite{abe2022deep} show that an ensemble's performance on out-of-distribution data depends on its performance on in-distribution data, and a larger single model can replicate the major benefits.

Many works have explored ways to give diverse predictions. Annealed Stein's Variational Gradient Descent provides a way to avoid the mode collapse by using a cyclic annealing for SVGD (\cite{d2021annealed}). It is based on rescaling the target distribution by tempering the posterior. The annealing aims to provide a well-explored initialization, and a linear schedule from [0,1] does not work well because of slow dynamics. To better trade-off exploration and convergence, they use cyclic hyperbolic annealing. Experiments for ERGD and s-ERGD use a linear schedule for $\beta$ between 1 and a positive constant, which works well for the entire training. Stein Variational Policy Gradient Descent has utilized a similar approach for regularizing the policy objective by tempering the target posterior.

In the context of POVI, making diverse predictions in weight-space does not mitigate the issue of non-identifiability in large networks, and function-space methods as described in \cite{d2021repulsive} and \cite{sun2019functional} perform better.

\section{Conclusion}
This paper provides a simple method for dealing with the issue of mode collapse in particle-based optimization methods. It highlights how having a symmetric kernel to model the interaction between the particles can be used to control the diversity in evolving particles. It proposes a way to train an ensemble by regularizing against this interaction. The work establishes the use of the method for classification tasks and shows how it can be helpful for efficient training for uncertainty-aware dynamics models. This demonstrates a way to reduce the dependence of the ensemble diversity on initial distribution while considering the particle interaction during training. 

\subsection{Acknowledgments}
I would like to thank Amy Zhang for their guidance, insightful discussions and feedback that were instrumental in shaping the direction of this study. Suggestions they provided facilitated the development of the methodology and without their encouragement and advice, this work would not have been possible.

\bibliography{paper}

\newpage

\onecolumn

\title{ENTROPY-REGULARIZED GRADIENT ESTIMATORS
FOR APPROXIMATE BAYESIAN INFERENCE\\(Supplementary Material)}
\maketitle
\appendix
\section{Appendix}
\subsection{The gradient flow of ERGD}
In the following section, it can be shown that ERGD minimizes the KL Divergence between the target distribution, and its approximation and CrossEntropy between the approximation and the kernel.
As described in \cite{d2021repulsive}, the particle evolution is described by the following equation. 
\begin{equation}
\begin{split}
      \frac{dx_{i}}{dt} = \frac{1}{n} \sum_{j=1}^n \left( k(\mathbf{x}_i, \mathbf{x}_j) \nabla_{\mathbf{x}_j} \log \pi (\mathbf{x}_j) - \beta \nabla_{\mathbf{x}_i} k(\mathbf{x}_i, \mathbf{x}_j) \right)
\end{split}
\end{equation}
\begin{equation}
 \begin{split}
    \frac{dx}{dt} = \int \left ( k(\mathbf{x},\mathbf{x'}) \nabla_{\mathbf{x'}} \log \pi (\mathbf{x'}) + 
   \nabla_{\mathbf{x'}} k(\mathbf{x},\mathbf{x'}) \right ) \rho(x') dx' + \int \left( - \nabla_{\mathbf{x'}} k(\mathbf{x},\mathbf{x'})  - \beta
   \nabla_{\mathbf{x}} k(\mathbf{x},\mathbf{x'}) \right ) \rho(x') dx' 
\end{split}
\end{equation}
Because of the symmetry of the kernel, the equation can be written as :
\begin{equation}
 \begin{split}
    \frac{dx}{dt} = \int \left ( k(\mathbf{x},\mathbf{x'}) \nabla_{\mathbf{x'}} \log \pi (\mathbf{x'}) + 
   \nabla_{\mathbf{x'}} k(\mathbf{x},\mathbf{x'}) \right ) \rho(x') dx' + \int \left( - \nabla_{\mathbf{x'}} k(\mathbf{x},\mathbf{x'})  + \beta
   \nabla_{\mathbf{x'}} k(\mathbf{x},\mathbf{x'}) \right ) \rho(x') dx' 
\end{split}
\end{equation}
\begin{equation}
 \begin{split}
    \frac{dx}{dt} = \int \left ( k(\mathbf{x},\mathbf{x'}) \nabla_{\mathbf{x'}} \log \pi (\mathbf{x'}) + 
   \nabla_{\mathbf{x'}} k(\mathbf{x},\mathbf{x'}) \right ) \rho(x') dx' + \int \left ( \left(  -1 + \beta  \right )
   \nabla_{\mathbf{x'}} k(\mathbf{x},\mathbf{x'}) \right ) \rho(x') dx' 
\end{split}
\end{equation}

\begin{equation}
 \begin{split}
    \frac{dx}{dt} = \int \left ( k(\mathbf{x},\mathbf{x'}) \nabla_{\mathbf{x'}} \log \pi (\mathbf{x'}) + 
   \nabla_{\mathbf{x'}} k(\mathbf{x},\mathbf{x'})  \rho(x') dx' + \int \left(  -1 + \beta  \right )
   \nabla_{\mathbf{x'}} \log k(\mathbf{x},\mathbf{x'}) \right ) \rho(x') dx' 
\end{split}
\end{equation}
\begin{equation}
\begin{split}
    \frac{dx}{dt} =  \int  k(\mathbf{x},\mathbf{x'}) \nabla_{\mathbf{x'}} \log \pi (\mathbf{x'}) \rho(x') dx' + 
   \nabla_{\mathbf{x'}} k(\mathbf{x},\mathbf{x'}) \rho(x') dx' - k(\mathbf{x},\mathbf{x'}) \rho(x') dx'  + \\
   \int \left ( \left( -1 + \beta  \right )
   \nabla_{\mathbf{x'}} \log k(\mathbf{x},\mathbf{x'}) \right ) \rho(x') dx' 
\end{split}
\end{equation}
\begin{equation}
\begin{split}
    \frac{dx}{dt} =  \int  k(\mathbf{x},\mathbf{x'}) \nabla_{\mathbf{x'}} \log \pi (\mathbf{x'}) \rho(x') dx' - 
   \int k(\mathbf{x},\mathbf{x'}) \nabla_{\mathbf{x'}} \rho(x') dx'  + \\
   \int \left ( \left( -1 + \beta  \right )
   \nabla_{\mathbf{x'}} \log k(\mathbf{x},\mathbf{x'}) \right ) \rho(x') dx' 
\end{split}
\end{equation}
\begin{equation}
\begin{split}
    \frac{dx}{dt} =  \int  k(\mathbf{x},\mathbf{x'}) \nabla_{\mathbf{x'}} \log \pi (\mathbf{x'}) \rho(x') dx' - 
   \int k(\mathbf{x},\mathbf{x'}) \nabla_{\mathbf{x'}} \log \rho(x') \rho(x') dx'  + \\
   \int \left ( \left( -1 + \beta  \right )
   \nabla_{\mathbf{x'}} \log k(\mathbf{x},\mathbf{x'}) \right ) \rho(x') dx' 
\end{split}
\end{equation}
\begin{equation}
\begin{split}
    \frac{dx}{dt} =  \int  k(\mathbf{x},\mathbf{x'}) \left ( \nabla_{\mathbf{x'}} \log \pi (\mathbf{x'}) - \nabla_{\mathbf{x'}} \log \rho(x') \right ) \rho(x') dx'  + \\
   \int \left ( \left( -1 + \beta  \right )
   \nabla_{\mathbf{x'}} \log k(\mathbf{x},\mathbf{x'}) \right ) \rho(x') dx' 
\end{split}
\end{equation}
\begin{equation}
 \begin{split}
   \frac{dx}{dt} = - \int \left ( k(\mathbf{x},\mathbf{x'}) \left [ \nabla_{\mathbf{x'}}  ( \left( -1 + \beta  \right ) \frac{\delta}{\delta \rho} H(\rho, \kappa) 
   + \nabla_{\mathbf{x'}} \frac{\delta}{\delta \rho} D_{KL}(\rho, \pi) \right ] \rho(x')  dx' \right )
\end{split}
\end{equation}
\begin{equation}
 \begin{split}
   \frac{dx}{dt} = - \int \left ( k(\mathbf{x},\mathbf{x'}) \left [ \nabla_{\mathbf{x'}} \frac{\delta}{\delta \rho} \left ( \left( -1 + \beta  \right )  H(\rho, \kappa) 
   +  D_{KL}(\rho, \pi) \right ) \right ] \rho(x')  dx' \right )
\end{split}
\end{equation}

\section{Experimental Details}

\subsection{Classification Tasks} For the CIFAR-10 classification task, ResNet-20 was trained over 50,000 epochs using a learning rate of 0.00025 across all methods, with 10 particles. Similarly, for FMNIST, ResNet-8 was trained for 50,000 epochs with a learning rate of 0.001 and an ensemble size of 10. A prior variance of 0.1 was used.

\renewcommand{\arraystretch}{1.5}
\begin{table}[h]
    \centering
    \caption{Hyperparameters for CIFAR-10 Classification}
    \begin{tabular}{|c|c|c|}
        \hline
        \textbf{Parameter} & \textbf{Value} & \textbf{Description} \\ 
         \hline
        Ensemble Size      & 10         & Number of Networks in the Ensemble \\ 
        \hline
        Learning Rate      & 0.00025         & Controls the step size during optimization \\ 
        \hline
        Batch Size         & 128             & Number of samples per batch \\ 
        \hline
        beta               & 1.6, 1.1          & Entropy parameter for ERGD, sERGD \\ 
        \hline
        Epochs             & 50,000          & Number of training iterations \\ 
        \hline
        Prior              & Normal (0,0.1)   & Prior for Network Parameters \\ 
        \hline
        Model Architecture & ResNet-20       & Type of Model used \\
        \hline
        Random Seeds       & 77, 89, 65, 42, 23  & Random Seeds for experiments \\ 
        \hline
    \end{tabular}
    \label{table:hyperparameters1}
\end{table}

\renewcommand{\arraystretch}{1.5}
\begin{table}[h]
    \centering
    \caption{Hyperparameters for FMNIST Classification}
    \begin{tabular}{|c|c|c|}
        \hline
        \textbf{Parameter} & \textbf{Value} & \textbf{Description} \\ 
         \hline
        Ensemble Size      & 20         & Number of Networks in the Ensemble \\ 
        \hline
        Learning Rate      & 0.001         & Controls the step size during optimization \\ 
        \hline
        Batch Size         & 256             & Number of samples per batch \\ 
        \hline
        beta               & 1.6, 1.1          & Entropy parameter for ERGD, sERGD \\ 
        \hline
        Epochs             & 50,000          & Number of training iterations \\ 
        \hline
        Prior              & Normal (0,0.1)   & Prior for Network Parameters \\ 
        \hline
        Model Architecture &  3 Layers, 128 units       & Type of Model used \\
        \hline
        Random Seeds       & 77, 89, 65, 42, 23  & Random Seeds for experiments \\ 
        \hline
    \end{tabular}
    \label{table:hyperparameters2}
\end{table}

\newpage
\subsection{Model-Based RL} 

\renewcommand{\arraystretch}{1.5}
\begin{table}[h]
    \centering
    \caption{Hyperparameters for HalfCheetah}
    \begin{tabular}{|c|c|c|}
        \hline
        \textbf{Parameter} & \textbf{Value} & \textbf{Description} \\ 
        \hline
        \textbf{Ensemble Size}      & 5              & Number of Networks in the Ensemble \\ 
        \hline
        \textbf{Number of Trials}        & 400         & Number of Episodes \\ 
        \hline
        \textbf{Trial Length}       & 1000           & Length of each trial \\ 
        \hline
        \textbf{Seed}               & 98,78,42             & Random seed value for experiments \\ 
        \hline
        \textbf{Training Hidden Layers} & 3         & Number of hidden layers in model \\ 
        \hline
        \textbf{Training Hidden Size}   & 256         & Number of neurons in each hidden layer \\ 
        \hline
        \textbf{Training Batch Size}   & 256         & Number of samples per batch \\ 
        \hline
        \textbf{Training Beta}        & 1.6             & Beta parameter for training \\ 
        \hline
        \textbf{Training Num Particles} & 20         & Number of particles for training \\ 
        \hline
        \textbf{Training Epochs}     & 100           & Number of training epochs \\ 
        \hline
        \textbf{Training Learning Rate} & 0.001      & Learning rate for training \\ 
        \hline
        \textbf{Optimizer Iterations} & 5            & Number of iterations for optimizer \\ 
        \hline
        \textbf{Optimizer Alpha}     & 0.1           & Alpha parameter for optimizer \\ 
        \hline
        \textbf{Optimizer Population Size} & 800     & Population size for optimizer \\ 
        \hline
        \textbf{Optimizer Planning Horizon} & 30     & Planning horizon for optimizer \\ 
        \hline
    \end{tabular}
    \label{table:hyperparameters3}
\end{table}

\renewcommand{\arraystretch}{1.5}
\begin{table}[h]
    \centering
    \caption{Hyperparameters for Pusher}
    \begin{tabular}{|c|c|c|}
        \hline
        \textbf{Parameter} & \textbf{Value} & \textbf{Description} \\ 
        \hline
        \textbf{Ensemble Size}      & 5              & Number of Networks in the Ensemble \\ 
        \hline
        \textbf{Number of Trials}        & 50         & Number of Episodes \\ 
        \hline
        \textbf{Trial Length}       & 100           & Length of each trial \\ 
        \hline
        \textbf{Seed}               & 98,78,42             & Random seed value for experiments \\ 
        \hline
        \textbf{Training Hidden Layers} & 3         & Number of hidden layers in model \\ 
        \hline
        \textbf{Training Hidden Size}   & 500         & Number of neurons in each hidden layer \\ 
        \hline
        \textbf{Training Batch Size}   & 32         & Number of samples per batch \\ 
        \hline
        \textbf{Training Beta}        & 1.6             & Beta parameter for training \\ 
        \hline
        \textbf{Training Num Particles} & 20         & Number of particles for training \\ 
        \hline
        \textbf{Training Epochs}     & 200           & Number of training epochs \\ 
        \hline
        \textbf{Training Learning Rate} & 0.001      & Learning rate for training \\ 
        \hline
        \textbf{Optimizer Iterations} & 5            & Number of iterations for optimizer \\ 
        \hline
        \textbf{Optimizer Alpha}     & 0.1           & Alpha parameter for optimizer \\ 
        \hline
        \textbf{Optimizer Planning Horizon} & 25     & Planning horizon for optimizer \\ 
        \hline
    \end{tabular}
    \label{table:hyperparameters4}
\end{table}

\renewcommand{\arraystretch}{1.5}
\begin{table}[h]
    \centering
    \caption{Hyperparameters for Cartpole}
    \begin{tabular}{|c|c|c|}
        \hline
        \textbf{Parameter} & \textbf{Value} & \textbf{Description} \\ 
        \hline
        \textbf{Ensemble Size}      & 5              & Number of Networks in the Ensemble \\ 
        \hline
        \textbf{Number of Trials}        & 50         & Number of Episodes \\ 
        \hline
        \textbf{Trial Length}       & 200           & Length of each trial \\ 
        \hline
        \textbf{Seed}               & 98,78,42             & Random seed value for experiments \\ 
        \hline
        \textbf{Training Hidden Layers} & 3         & Number of hidden layers in model \\ 
        \hline
        \textbf{Training Hidden Size}   & 256         & Number of neurons in each hidden layer \\ 
        \hline
        \textbf{Training Batch Size}   & 256         & Number of samples per batch \\ 
        \hline
        \textbf{Training Beta}        & 1.6             & Beta parameter for training \\ 
        \hline
        \textbf{Training Num Particles} & 20         & Number of particles for training \\ 
        \hline
        \textbf{Training Epochs}     & 200           & Number of training epochs \\ 
        \hline
        \textbf{Training Learning Rate} & 0.001      & Learning rate for training \\ 
        \hline
        \textbf{Optimizer Iterations} & 5            & Number of iterations for optimizer \\ 
        \hline
        \textbf{Optimizer Alpha}     & 0.1           & Alpha parameter for optimizer \\ 
        \hline
        \textbf{Optimizer Planning Horizon} & 25     & Planning horizon for optimizer \\ 
        \hline
    \end{tabular}
    \label{table:hyperparameters5}
\end{table}

\end{document}